\pdfoutput=1
\documentclass[12pt,a4paper]{article}
\usepackage[T1]{fontenc}
\usepackage{lmodern}
\usepackage[english]{babel}
\usepackage{graphicx}
\usepackage{caption}
\usepackage{amsmath}
\usepackage[toc]{appendix}
\usepackage{amsfonts}
\usepackage[square, numbers]{natbib}
\usepackage{xcolor}
\usepackage{CJKutf8}
\usepackage{booktabs}
\usepackage{colortbl}
\usepackage{array}
\usepackage{makecell}
\usepackage{authblk}
\usepackage{listings}
\usepackage[colorlinks, citecolor=blue]{hyperref}
\usepackage{setspace}

\title{\LARGE{Textual Similarity as a Key Metric in Machine Translation Quality Estimation}}
\date{\vspace{10ex}}

\author[1]{\footnotesize Kun Sun\thanks{\texttt{kun.sun@uni-tuebingen.de}}}
\author[2]{\footnotesize Rong Wang}
\affil[1]{\footnotesize Department of Linguistics, University of Tübingen, Germany}
\affil[2]{\footnotesize Institute of Natural Language Processing, Stuttgart University, Stuttgart, Germany}

\begin{document}

\maketitle
\vspace{0.5cm}
\begin{abstract}
Machine Translation (MT) Quality Estimation (QE) assesses translation reliability without reference texts. This study introduces ``textual similarity'' as a new metric for QE, using sentence transformers and cosine similarity to measure semantic closeness. Analyzing data from the MLQE-PE dataset, we found that textual similarity exhibits stronger correlations with human scores than traditional metrics (hter, model evaluation, sentence probability etc.). Employing GAMMs as a statistical tool, we demonstrated that textual similarity consistently outperforms other metrics across multiple language pairs in predicting human scores. We also found that ``hter'' actually failed to predict human scores in QE. Our findings highlight the effectiveness of textual similarity as a robust QE metric, recommending its integration with other metrics into QE frameworks and MT system training for improved accuracy and usability.

\end{abstract}

\small{{\bf Keywords:} {translation quality evaluation, sentence transformers, cosine similarity, regression analysis, key metric}

\clearpage

\section{Introduction}

There are two kinds of quality evaluations in machine translation (MT): ``MT quality evaluation'' and ``MT quality estimation'' \cite{specia2010machine}. Each field has distinct standards and applications, ensuring comprehensive translation quality assessment.  The former is used to compare the quality of translated texts when the reference translations are available. In contrast, the latter, also named as MT ``quality estimation (QE)'', involves assessing the quality of MT outputs without reference translations \cite{ranasinghe2020transquest}. QE is essential for various applications, such as determining whether an automatically translated sentence or document is ready for the end user or requires human post-editing. It can flag passages with critical errors, serve as a quality metric when reference translations are unavailable, and assist in computer-aided translation interfaces by highlighting text needing human revision and estimating the required human effort. The current study focuses on the latter, quality estimation.

``MT quality evaluation'' involves using various automatic metrics to assess translation accuracy and effort. Common metrics include BLEU, which measures n-gram overlap with reference translations and applies a brevity penalty; METEOR, which considers precision, recall, synonyms, and stemming; TER, which calculates the number of edits needed to match a reference translation; chrF, which evaluates character n-grams, useful for morphologically rich languages; and BERTScore, which uses pre-trained embeddings to assess semantic similarity. Moreover, supervised metrics, trained using human-judged data often show a higher correlation with human evaluations, and some metrics were used such as BEER, BLEND \cite{lee2023survey}.

However, in MT QE, several features play pivotal roles in both quality estimation and enhancing the overall translation process. The ``model eveluation'' (ML\_eval) quantifies the MT model's confidence in its translations. This metric is typically derived from the log-likelihood of the translation given the source sentence. 
Higher scores reflect greater confidence, indicating that the translation aligns well with the model's learned patterns from the training data. These scores are crucial for quality estimation TM systems, helping to predict the reliability of translations from TM models and highlight areas that may require further human intervention or correction \cite{blain2023findings}, \cite{fomicheva2022mlqe}. Moreover, in many cases, sentences with higher probabilities from n-gram models do tend to be more natural-sounding. This is because these models are trained on large corpora of text, capturing common patterns in language use. Sentence probabilities for translated texts are also used to evaluate translation quality without reference translations.

Another significant feature is ``human translation edit rate'' (hter) \cite{snover2006study}, and it measures the number of edits required to correct a translation output from MT to match a human reference. It accounts for insertions, deletions, substitutions, and shifts, providing a quantitative measure of the effort needed for post-editing. This makes \texttt{hter} a practical metric for assessing the quality of MT systems, as it directly correlates with the human effort required to produce accurate translations. 


In MT quality estimation, ``ML\_eval'' often serve as one key feature in helping to gauge the reliability and quality of TM systems when the reference translations are not available. The metric of ``ML\_eval'', combined with linguistic features and contextual embeddings, enhance the accuracy of TM quality assessments. They help identify translations that may require human review or post-editing. Further, in the training or fine-tuning of NMT (neural machine translation) or LLMs-based TM models, the metric of ``ML\_eval'' still plays an important role, such as on optimizing translation performance by leveraging large parallel corpora, word alignments, and contextual learning techniques \cite{guzman2019flores}, \cite{ott2019fairseq}. Even if the metric is not directly used in training or fine-tuning, it still could provide valuable feedback on the translation quality, guiding iterative improvements and fine-tuning of the model. This distinction underscores the dual role of ``ML\_eval'' in both evaluating and enhancing translation quality. 

While ``ML\_eval'' provides valuable insights into QE, the metric has certain limitations. One significant issue is the tendency to overestimate confidence in poor translations, often due to overfitting or a lack of diverse training data. This overconfidence can lead to inaccuracies in QE. Moreover, the metric is sensitive to the training data distribution, and meaning biases or gaps in the data can skew the scores. Third, this metric may also fail to capture contextual meaning and subtle semantic differences that human evaluators can detect, leading to discrepancies between model scores and perceived translation quality. Additionally, although higher probability from an n-gram model often correlates with more natural-sounding sentences, this metric may not be a ideal indicator.  The reason for this is that the relationship between statistical probability and perceived naturalness is complex and influenced by many factors beyond  n-gram probabilities.


The metric, ``hter'',  has also some weaknesses. First, the variability in human editing styles and preferences can lead to inconsistent hter scores, making it challenging to standardize quality assessments. Additionally, ``hter'' focuses on the number of edits rather than the nature of the changes, failing to differentiate between minor stylistic tweaks and substantial corrections. This can skew the perception of translation quality. Furthermore, the metric relies on the availability of high-quality reference translations, which may not always be accessible, limiting its applicability in certain contexts. These limitations suggest that while ``hter'' is useful, it should be complemented with other evaluation methods for a more comprehensive assessment of translation quality.

We have identified some limitations in the two key metrics used in Quality Estimation (QE). Despite these, traditional QE methods face a significant, often overlooked problem: the reliance on \textbf{correlation} statistical analysis for evaluation. Correlation is primarily used to understand the relationship between two data sets. Correlation only indicates the degree to which two sets of data are related, but it does not establish a relationship between them. However, to determine the effect of one variable on another, regression analysis is essential. For a deeper insight into key features affecting machine learning quality estimation data, regression analysis provides a more robust approach. For example, if we want to determine whether ``hter'' has an effect on human scores, regression analysis is necessary. The correlation between ``hter'' and human scores does not provide this level of insight.


To address these weaknesses, it is crucial to complement existing metrics with additional features, continuously update and diversify training data, and incorporate new features to enhance the reliability and usability of QE. Upgrading statistical analysis methods is also necessary to gain deeper insights. This study proposes that ``textual similarity'' could be considered a key feature in MT quality estimation and training. Using existing MT quality estimation datasets, we employed advanced statistical methods to compare this new metric with existing features and explore their advantages. 
 
\section{Methods}

\subsection{Dataset}

This study utilized two distinct datasets. The first dataset, \texttt{MLQE-PE} \cite{fomicheva2022mlqe}, and the second dataset, used in \texttt{PreQuEL} \cite{don2022prequel}. Although the datasets differ in size, they can both be effectively employed for cross-validation purposes.

\texttt{MLQE-PE} is a comprehensive dataset for MT QE and Automatic Post-Editing, covering eleven language pairs, including both high- and low-resource languages \cite{fomicheva2022mlqe}. The dataset features up to 10,000 translations per language pair, annotated with sentence-level direct assessments, post-editing effort, and word-level binary good/bad labels. Each source-translation pair includes the post-edited sentence, article titles, and details of the neural MT models used. The dataset is thoroughly documented and analyzed, and the information on the baseline system performances is included.

The main variables in the \texttt{MLQE-PE} includes: 
original (original sentence), translation (MT output),
scores (list of DA [direct assessment] scores by all annotators - the number of annotators may vary),  mean (average of DA scores), 
z\_scores (list of z-standardized DA scores), z\_mean (average of z-standardized DA scores), model\_scores (NMT model score for sentence). The other information on sentence translation such as ``hter'' (human translation edit rate) is also included in the dataset. Note that ``model\_scores'' is the  the MT model’s confidence in its translation. According to MLQE-PE, these MT systems are SOTA ones, representing the recent advancements in MT models. In this way, ``model\_scores'' signifies a direct indication of the translation's reliability from MT, that is, the translation quality assessment. The higher scores represents the higher translation quality. Conversely, a lower ``hter'' indicates that a translation needs more human efforts to post-edit, that is, the translation quality is higher. Moreover, we also created some factors. For instance, the standard variation (sd) of different human scores for the same translation, and the human annotator number. 

There are 11 language pairs available: English-German (en-de), English-Chinese (en-zh), Romanian-English (ro-en), Estonian-English (et-en), Nepalese-English (ne-en), Sinhala-English (si-en), and Russian-English (ru-en). \footnote{Although the paper on MLQE-PE claimed that their dataset also includes other language pairs, the dataset on these language pairs in the \texttt{github} they provided is not available.}

\texttt{PreQuEL} \cite{don2022prequel} includes a variety of datasets. However, we selected only the datasets for evaluating machine translation quality without reference translations. In this case, there are two language pairs: English-Chinese and English-German. The dataset includes the following variables: n-gram sentence probability, language model score, HTER, and human score (z\_mean). ``Sentence probability'' refers to the likelihood of a particular sequence of words (sentence) occurring. This can be computed using n-grams. For example, if bi-grams are used, it is termed ``bi-gram sentence probability'', and if tri-grams are used, it is ``tri-gram sentence probability''.  \texttt{PreQuEL} provides five types of sentence probabilities using 1-5 grams.
The ``language model score'' is a metric related to the language probability score (``lan'' is indicated in the \texttt{PreQuEL}). This score is part of the quality estimation process that evaluates how likely a translation is accurate based on linguistic features and models, without referencing the actual translated text. \texttt{PreQuEL} also provides ``hter'', which is computed using the same method as in \texttt{MLQE-PE}. Additionally, \texttt{PreQuEL} provides only the ``human score (z\_mean)''.

\subsection{Textual similarity}

Textual similarity (or sentence similarity) measures how similar or different two pieces of text are semantically. This can involve comparing sentences, paragraphs, or entire documents to see how closely they match in meaning or content. It can be calculated using various techniques, such as comparing word overlaps, semantic similarity, or using machine learning models to assess how alike the texts are. However, after transformers revolutionized deep learning, textual similarity can now be effectively measured using existing language models. \cite{vaswani2017attention}, \cite{devlin2018bert}, \cite{wolf2020transformers}

Pretrained \texttt{sentence transformers} are highly effective for generating sentence or textual embeddings due to their ability to capture deep semantic meanings \cite{reimers2019sentence}. By leveraging the Transformer architecture, these models encode sentences into high-dimensional vectors that reflect their contextual and semantic content, rather than merely their lexical features. This allows for accurate comparisons of semantic similarity between sentences or texts, which is crucial for tasks like paraphrase detection, information retrieval, and clustering.

The ``sentence-transformers/paraphrase-multilingual-MiniLM-L12-v2'' model creates dense vector representations of sentences, capturing the semantic essence of the text. These vectors are placed in a shared embedding space where similar meanings are mapped to nearby points. The model's fine-tuning for tasks like paraphrase identification ensures that sentences with similar meanings are close in this space. 
The use of sentence transformers is particularly valuable in multilingual contexts, as they support multiple languages and capture semantic nuances effectively. The task in the current study is to understand the source text in one language and the translated texts in the other language. The capability of multilingual sentence transformer models is essential for applications to explore the semantic similarity among languages.  The current study employs this multilingual sentence transformer model to compute semantic similarity for one source sentence and its translation. Our fundamental approach involves using the sentence transformer to generate text embeddings and then calculating their similarity using the cosine method. We applied the method to process the two datasets to compute ``textual similarity'' between a source text and its translated text. 


\subsection{Regression statistical analysis}
We applied Generalized Additive Mixed Models (\textbf{GAMM}) in analyzing how factors influence human scores. GAMMs incorporate non-linear relationships between the dependent and independent variables through smooth functions \citep{wood2017generalized}. GAMMs allow for both fixed and random effects, accommodating complex variations within hierarchical data structures. The ``additive'' part of GAMM means that the model expresses the dependent variable as a sum of smooth functions of predictors, along with any random effects and an error term. This flexibility makes GAMMs particularly useful for modeling non-linear trends in data, where the effect of variables is not strictly linear and may vary by group or over time.

GAMM could leverage the function \texttt{s()}. This smooth function better gets model fittings for some factors, and the interaction smooth could find the interaction among some given factors. Some random variables could play a very important role, such as different language pairs in TM. The role of such random variables could be well explored by using GAMMs. GAMMs are friendly to make model comparison by referring to \texttt{AIC} (Akaike information criterion is an estimator of prediction error). 

In GAMM setups, the independent variable is ``human score'', and other metrics are dependent variables. Some random factors, such as human evaluator number, different language pairs (source language - target language), could play an essential role \cite{barr2013random}. A GAMM fitting should include a number of factors. The reason for this is that various metrics or factor could predict human score or take effect on human score, and these factors co-work to play a role. The purpose of using GAMM fitting is to better explore how these factors take effects on human scores. 

\section{Results}
\subsection{Result 1: Correlations}
We plotted the Pearson correlations among various factors for \texttt{MLQE-PE}, as shown in Fig.~\ref{fig:corr}. The results indicate that ``ML\_eval'' is correlated with the human score (mean) at 0.15, with ``hter'' at 0.06, and with the human score (z-mean) at 0.3. ``Textual similarity'' shows a correlation of 0.47 with the human score (mean), -0.06 with the human score (z-mean), -0.45 with ``hter'', and -0.21 with ``ML\_eval''. Overall, ``textual similarity'' exhibits stronger correlations with other factors compared to ``ML\_eval''.

\begin{figure}[h]
    \includegraphics[width = 15cm]{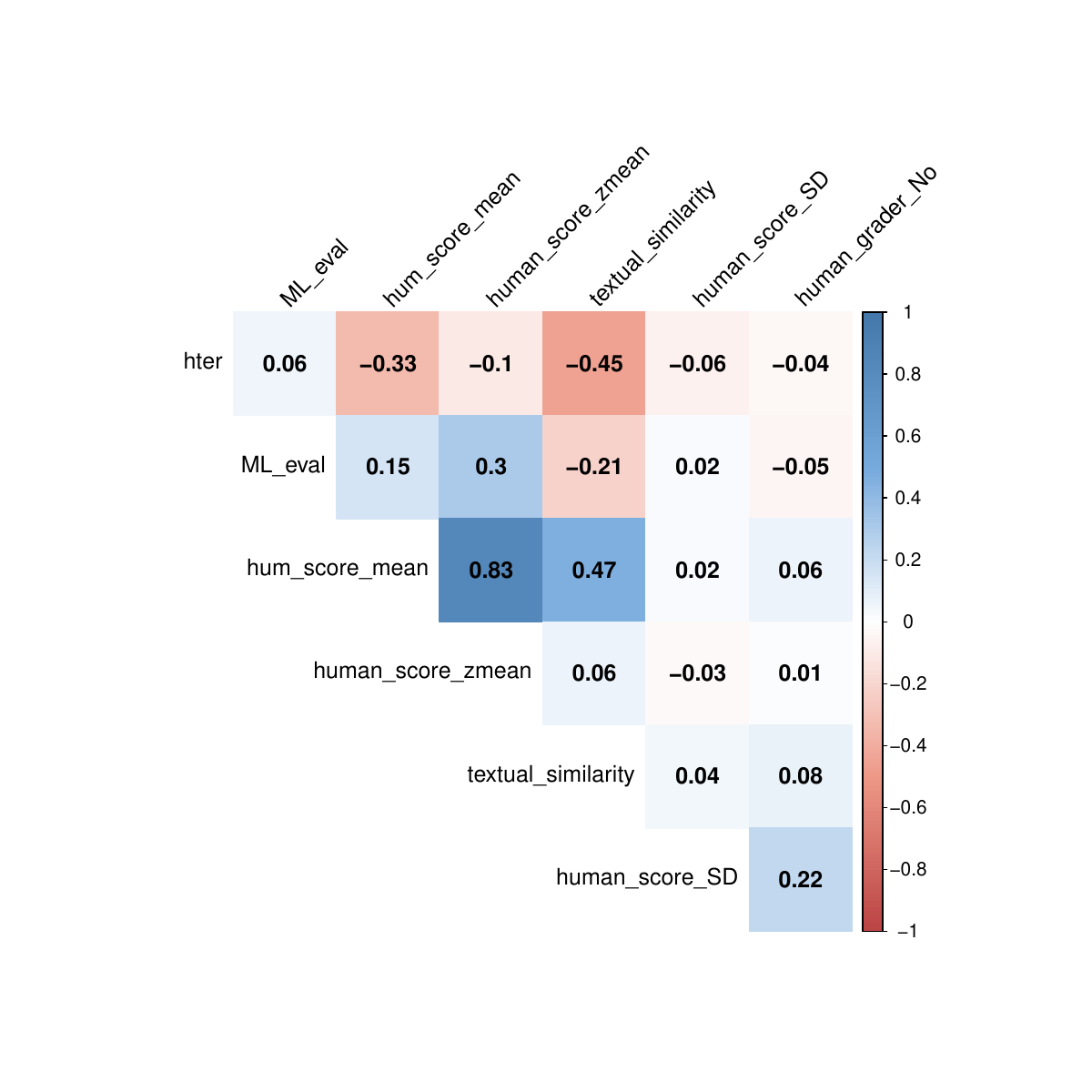}
    \caption{correlations among various factors in \texttt{MLQE-PE} }
    \label{fig:corr}
\end{figure}

Similarly, we plotted the correlation among various factors for \texttt{PreQuEL}, as shown in Fig~\ref{fig:prcorr} (see the Appendix). The three variables: ``n-gram sentence probability'', ``hter'', and ``textual similarity'', show similar correlation values with the ``human score (z\_mean)'', each approximately 0.15.

\subsection{Result 2: Cross language pairs}

 We established three types of GAMM fittings for \texttt{MLQE-PE}, where ``human score (mean)'' is the independent variable, and other factors include ``ML\_eval'', ``textual similarity'', ``sd of human scores''. ``human evaluator number'', ``language pairs (langs)'' are incorporated as random variables. We chose ``human score (mean)'' over ``human score (z-mean)'' as the independent variable primarily because ``human score (mean)'' closely follows a normal distribution. The GAMM setups are detailed below, and the results are presented in Table \ref{tab1}.

\begin{itemize}

 \item base model = bam(human score $\sim$ s(ML\_eval)+s(textual similarity)+s(sd)+\\s(hter)+s(evaluator num, bs=``re'')+s(langs, bs=``re''))
 
\item m1=bam(human score  $\sim$ s(ML\_eval)+s(sd)+s(hter)+s(evaluator num, bs=``re'')\\+s(langs, bs=``re''))

\item  m2=bam(human score $\sim$ s(textual similarity)+s(sd)+s(hter)+s(evaluato num, bs=``re'')+s(langs, bs=``re''))

\item m3=bam(human score $\sim$ s(ML\_eval)+s(textual similarity)+s(sd)+s(evaluator num, bs=``re'')+s(langs, bs=``re''))
\end{itemize}

The base model demonstrates the best performance when all factors are included. However, each of the remaining models includes only a subset of these factors. The Akaike Information Criterion (AIC) is used to represent the performance of this GAMM fitting. The base model has the lowest AIC. When the AIC of model $m1$ is subtracted from the AIC of the base model, the resulting $\Delta$AIC indicates the contribution of ``textual similarity'', as ``m1'' does not include ``textual similarity'' compared to the base model. A smaller $\Delta$AIC also indicates better performance and greater contribution for a given factor. The results are shown in Table \ref{tab1}.
\begin{table}
\centering
\caption{$\Delta$AIC for different GAMM fittings for \texttt{MLQE-PE}. A smaller $\Delta$AIC indicates better performance (n=45886)}
\begin{tabular}{ | c | c |  }
 \hline
factor (contribution)  &     $\Delta$AIC \\ \hline\hline
(m1- base model) contribution of textual similarity  & 848.87 \\ \hline               
(m2- base model) contribution of ML\_eval &   8785.82  \\ \hline    
(m3-base model) contribution of  hter    &  586.3 \\ \hline
\end{tabular}
\label{tab1}
\end{table}

The contribution from ``textual similarity'' outperforms that from ``ML\_eval''. However, the contribution of ``hter'' shows the best performance overall. This trend holds for all language pairs. Despite this, it is important to examine the performance for individual language pairs. The next section explores the performance of these factors in each language pair.

Next, we used similar GAMM methods to explore how different metrics affect human scores for \texttt{PreQuEL}. In the \texttt{PreQuEL} dataset, ``human score (z\_mean)'' is the dependent variable \footnote{In this dataset, ``human score (mean)'' is not available.}, and other factors include ``n-gram sentence probability'', ``language model score'', ``hter'', and ``textual similarity''. In this dataset, ``language model score'' (``lm\_score'') has four values, so we treat it as a random factor, and ``language pairs'' (``langs'') are incorporated as a random variable. There are five types of ``n-gram sentence probability''; however, we chose the optimal one, ``trigram sentence probability'',"for the GAMM fittings. The GAMM setups are detailed below, and the results are presented in Table \ref{tab2}.

\begin{itemize}

 \item base model = bam(human score $\sim$ s(trigram sent prob)+s(textual \\ similarity)+s(hter) +s(lm\_score, bs=``re'')+s(langs, bs=``re''))
 
\item t1=bam(human score  $\sim$ s(trigram sent prob)+s(hter)+s(lm\_score, bs=``re'')\\+s(langs, bs=``re''))

\item  t2=bam(human score $\sim$ s(textual similarity)+s(hter)+s(lm\_score, bs=``re'')\\+s(langs, bs=``re''))

\item t3=bam(human score $\sim$ s(trigram sent prob)+s(textual similarity)\\+s(lm\_score, bs=``re'')+s(langs, bs=``re''))
\end{itemize}

\begin{table}
\centering
\caption{$\Delta$AIC for different GAMM fittings for \texttt{PreQuEL}. A smaller $\Delta$AIC indicates better performance (n=14706)}
\begin{tabular}{ | c | c |  }
 \hline
factor (contribution)  &     $\Delta$AIC \\ \hline\hline
(t1- base model) contribution of textual similarity  & 237.11 \\ \hline               
(t2- base model) contribution of trigram sentence probability &   232.16  \\ \hline    
(t3-base model) contribution of  hter    &  424.29 \\ \hline
\end{tabular}
\label{tab2}
\end{table}

The contribution of ``textual similarity'' outperforms that of ``hter''. However, the contribution of ``trigram sentence probability'' shows a similar performance to ``textual similarity'', with ``trigram sentence probability'' demonstrating the best performance overall. The next section examines the performance for individual language pairs.

\subsection{Result 3: Individual language pair}
Using a similar GAMM setup as the base model, we explore how these factors perform across seven language pairs for \texttt{MLQE-PE}. For each language pair, we established the same GAMM and plotted the partial effects for each factor of interest, as shown in Figure \ref{fig:pe}. When the \textit{p}-value is greater than 0.05, the plot is not significant. Within the same language pair, a smaller $\Delta$AIC indicates better performance. Clearly, all cases for ``ML\_eval'' are significant, and there is only one insignificant case for ``textual similarity''. In contrast, ``hter'' has only two significant cases, indicating five insignificant cases.

Compared with ``ML\_eval'', ``textual similarity'' outperforms in the cases of ``German-English'', "English-Chinese'', ``Romanian-English'', ``Russian-English'' and ``Sinhala-English''.  Generally, ``textual similarity'' demonstrates the best performance across individual language pairs.
 
\begin{figure}[h]
    \includegraphics[width = 16.6cm]{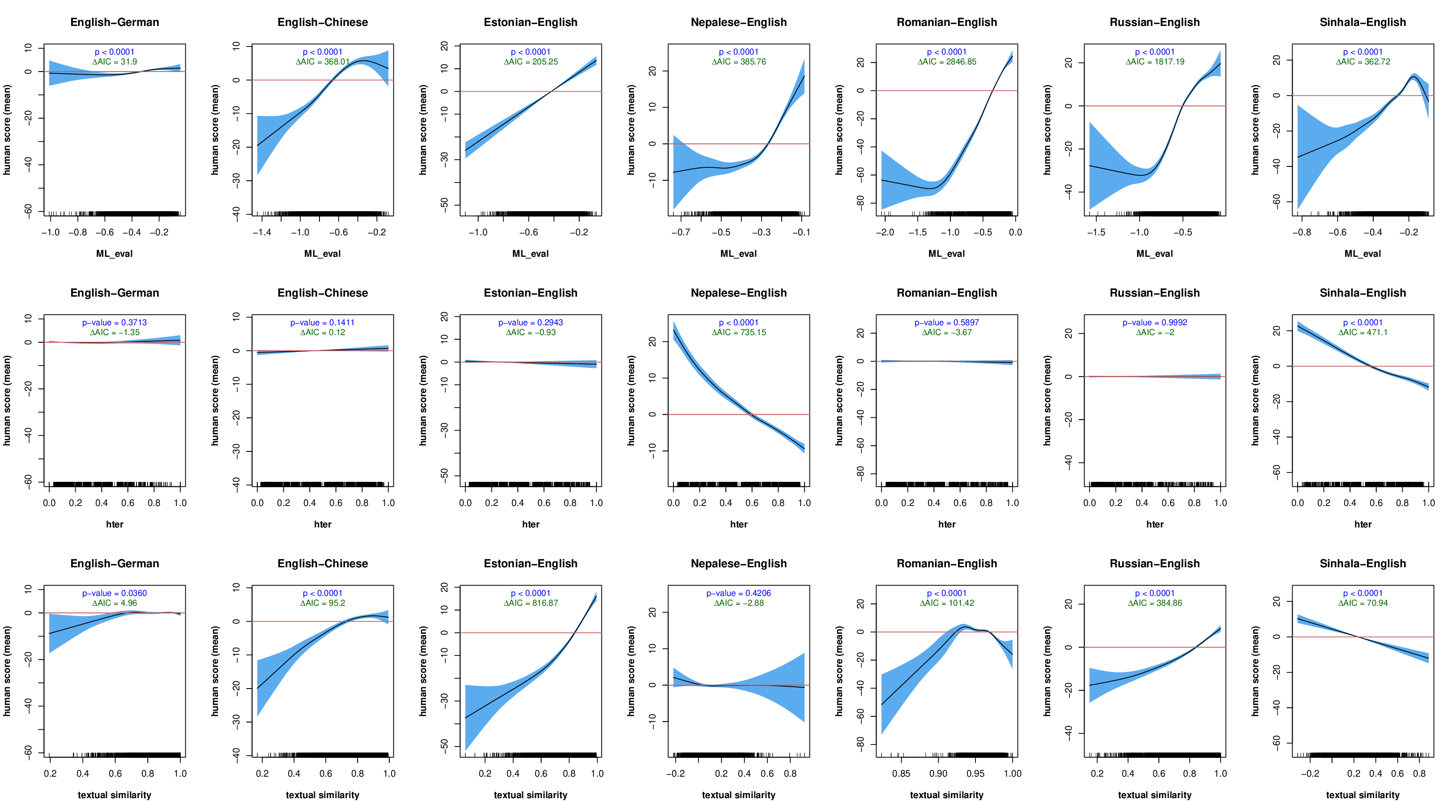}
    \caption{The partial effects on human score from different factors for  \texttt{MLQE-PE}. The \textit{x}-axis represents the specific metric being analyzed, while the \textit{y}-axis indicates human score. Each curve within a plot illustrates the relationship between a predictor variable (plotted on the \textit{x}-axis) and the response variable. Steeper slopes on these curves indicate a stronger influence of the predictor variable on human score. Conversely, gentler slopes imply a weaker influence, indicating that changes in the predictor variable have a less pronounced effect on human score. Such plots could give deep insights on the relationship between one given metric and human score.
}
    \label{fig:pe}
\end{figure}

Adopting the similar GAMM fittings for \texttt{PreQuEL}, we explored the performance for each variable of our interest in each language pair, and plotted the partial effects for each factor of interest, as shown in Figure \ref{fig:pr}. Figure \ref{fig:pr} shows that `` textual similariy'' has the best performance in each individual language pair.

\begin{figure}[h]
    \includegraphics[width = 12.6cm]{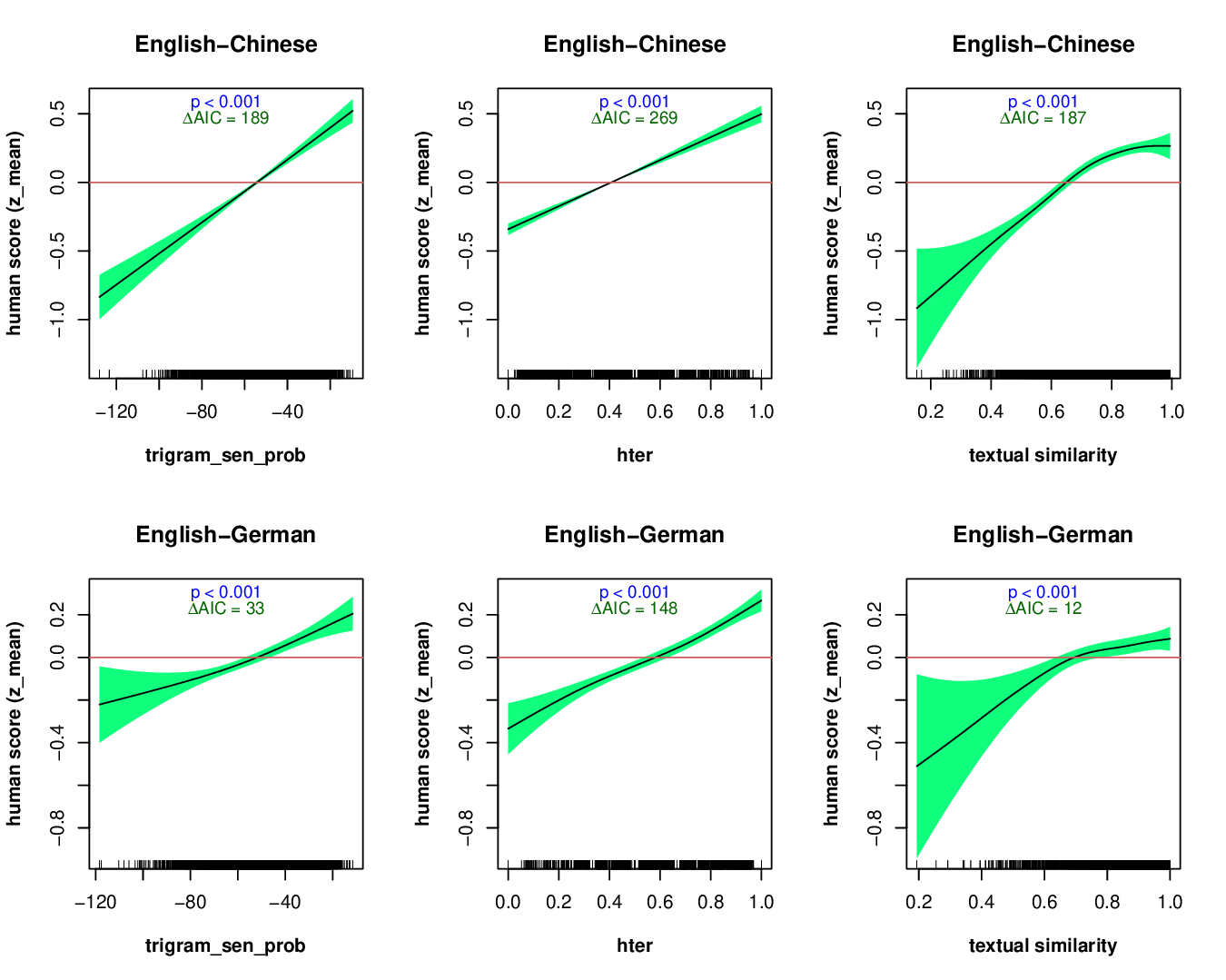}
    \caption{The partial effects on human score from different factors for \texttt{PreQuEL}. The \textit{x}-axis represents the specific metric being analyzed, while the \textit{y}-axis indicates the human score. The interpretation of the curve is the same as in Fig~\ref{fig:pe}. The layout here differs slightly from that in Fig~\ref{fig:pe}. $\Delta$AIC values are compared among three plots for the same language pairs. A lower $\Delta$AIC value indicates better performance.}
    \label{fig:pr}
\end{figure}

\section{Discussion}

Considering the correlation in \texttt{MLQE-PE}, we found that ``textual similarity'' exhibits a strong correlation with human scores, ``ML\_eval'', and ``hter''. In contrast, ``ML\_eval'' shows a weaker correlation with the human score (mean). When examining the GAMM fittings across various language pairs and in each case of language pair, the performance of ``textual similarity'' surpasses that of ``ML\_eval''. On an individual language pair basis, ``textual similarity'' also outperforms both ``ML\_eval'' and ``hter''.  In \texttt{PreQuEL}, due to the smaller sample size, the differences in correlation are not obvious.  In short, ``textual similarity'' as a metric could be closely correlated with human score, and is highly capable of predicting human score. 

The following provides a detailed analysis of each metric in \texttt{MLQE-PE}. ``ML\_eval'' demonstrates a significant impact across all language pairs, indicating that this metric is both useful and effective for QE. However, ``hter'' does not show a significant impact on human scores in the most cases, suggesting that this metric may not be suitable for evaluating MT quality in certain language pairs. On the other hand, ``textual similarity'' predicts human scores in the majority of language pairs, highlighting its potential as an effective metric for QE.

The present study underscores the importance of selecting appropriate metrics for QE. ``Textual similarity'', in particular, emerges as a robust and reliable metric that consistently correlates with human evaluation scores and effectively predict human scores, making it a valuable tool for improving the accuracy of QE. In contrast, while ``ML\_eval'' remains a useful metric, its effectiveness varies across different language pairs. ``hter'', despite being commonly used, may not be sufficient in some cases, necessitating the consideration of alternative or supplementary metrics like textual similarity.

Next, we analyzed the performance of ``textual similarity'' in \texttt{PreQuEL}. ``textual similarity'' outperforms ``hter'' consistently across language and within invidiual languages. However, ``textual similarity'' has the similar performance with ``n-gram sentence probability''. It also reveals that ``n-gram sentence probability'' may be a useful metric in evaluating MT QE. 

It is easy to understand why textual similarity is such an effective metric. When a translated text is semantically close to the source text, it indicates that the translation meets a crucial standard of quality: the meaning of the translation should closely match the original text. In contrast, the ``hter'' metric often fails for most language pairs because post-editing efforts do not necessarily reflect changes in meaning. For example, a translation may be very close in meaning to the original text but contain some misspellings. Humans may still consider the translation to be good, even though the post-edit rate is low due to the necessary corrections. Conversely, minor edits to core verbs or key words in a translation text may require minimal edit effort but significantly alter the meaning. However, such a translation text does not necessarily meet the standard of good translation. This discrepancy shows that post-edit rate does not always correlate with translation quality.

Therefore, exploring the cognitive recognition of what constitutes a good translation for different language users is worthwhile. Understanding this can significantly enhance both machine translation and quality estimation processes by aligning them more closely with human judgments of translation quality. For instance, "n-gram sentence probability" can provide an initial impression of the naturalness or readability of translated texts. If this metric is low, human evaluators might rate the quality of the translations poorly, regardless of how closely they match the original meaning. Simple yet effective metrics should be considered in evaluating machine translation quality, incorporating factors that align with human translation assessment standards and processes. This approach can lead to substantial progress in machine translation quality estimation (MT QE).

To date, ``textual similarity'' has not been proposed as a metric in QE \footnote{Although some studies have proposed using semantic similarity to evaluate MT quality, they have only applied this method in cases where reference translations are provided \cite{castillo2012semantic, zhang2019bertscore}. Specifically, these studies compared the semantic similarity between a candidate translation and a reference translation. In contrast, our study employed cross-lingual the semantic similarity between a text in the source language and its translated text in the target language.}. Our study applied a variety of statistical methods to analyze data from established QE datasets, demonstrating that this metric is both reliable and effective compared to commonly used metrics in QE. Our findings provide strong evidence that ``textual similarity'' is a robust metric, making it a valuable addition to the existing suite of QE metrics. Given its proven effectiveness, ``textual similarity'' should be included as a key metric in QE and incorporated into the training of machine translation (MT) systems. This integration can enhance the accuracy and reliability of MT quality assessments, ultimately improving the performance and usability of MT systems.

\bibliographystyle{plain}
\bibliography{reference}
\appendix
\section*{Appendix}

\begin{figure}[h]
    \includegraphics[width = 15cm]{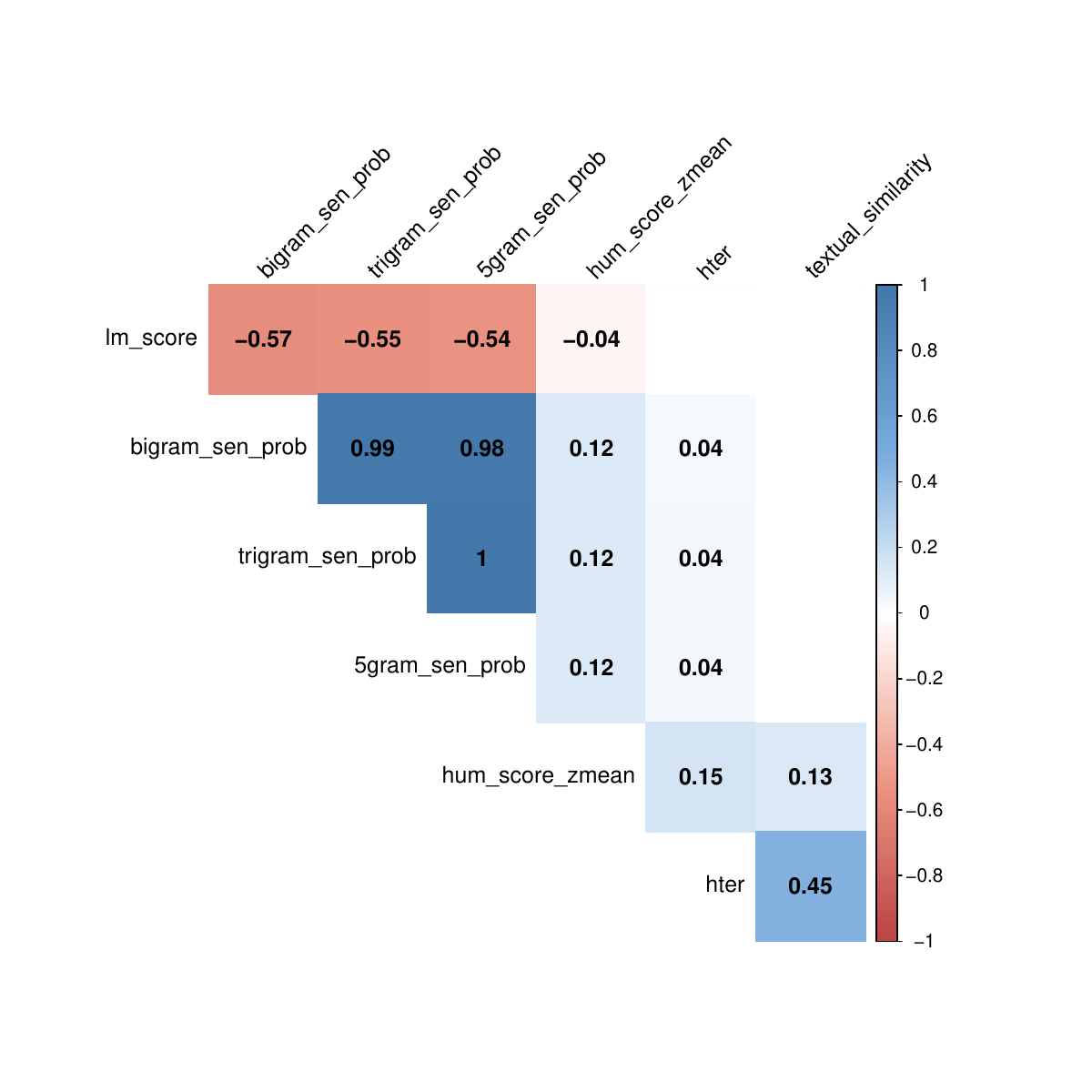}
    \caption{correlations among various factors in \texttt{PreQuEL} }
    \label{fig:prcorr}
\end{figure}
\end{document}